\documentclass[review,12pt]{elsarticle}

\usepackage{xr}
\usepackage{float}

\usepackage{amssymb}

\usepackage{booktabs}
\usepackage{siunitx}

\usepackage{graphicx}
\usepackage{graphics}

\usepackage{subfigure}
\usepackage{caption}
\usepackage[english]{babel}
\usepackage[utf8]{inputenc}
\usepackage[T1]{fontenc}

\usepackage{rotating}

\usepackage[ruled,,vlined,lined]{algorithm2e}

\usepackage{array}
\usepackage{array}
\usepackage{graphicx}
\usepackage{siunitx}

\usepackage{}
\usepackage{amsmath}
\usepackage{amssymb}
\usepackage{graphics}

\usepackage{subfigure}
\usepackage{lscape}
\usepackage{multirow, dcolumn, booktabs}
\usepackage{rotating}

\usepackage{url}

\newcommand{\red}{G=\left( V,E\right)}

\newtheorem{definition}{Definition}

\journal{Applied Soft Computing}

\begin{document}

\begin{frontmatter}

\title{Community Detection in Networks: A Rough Sets and Consensus Clustering Approach}

\author[inst1,inst2]{Darian H. Grass-Boada\corref{cor1}}
\author[inst1,inst2]{Leandro Gonz\'alez-Montesino}
\author[inst1,inst2]{Rub\'en Arma\~{n}anzas}

\affiliation[inst1]{organization={Institute of Data Science and Artificial Intelligence (DATAI)},
            addressline={Campus Universitario, Edificio Ismael S\'anchez Bella}, 
            city={Pamplona},
            postcode={31009}, 
            state={Navarra},
            country={Spain}}

\affiliation[inst2]{organization={TECNUN School of Engineering},
            addressline={Manuel Lardizabal Ibilbidea, 13}, 
            city={Donostia-San Sebastián, Gipuzkoa},
            postcode={20018}, 
            state={País Vasco},
            country={Spain}}

\cortext[cor1]{Correspondence to: dhgrass@unav.es (D.H. Grass-Boada)}

\begin{abstract} 
The objective of this paper is to propose a framework, called Rough Clustering-based Consensus Community Detection (RC-CCD), to effectively address the challenge of identifying community structures in complex networks from a set of different community partitions. The method uses a consensus approach based on Rough Set Theory (RST) to manage uncertainty and improve the reliability of community detection. The RC-CCD framework is tested on synthetic benchmark networks generated by the Lancichinetti–Fortunato–Radicchi (LFR) method, which simulate varying network scales, node degrees, and community sizes. Key findings demonstrate that RC-CCD outperforms established algorithms like Louvain, Greedy, and LPA in terms of normalized mutual information, showing superior accuracy and adaptability, particularly in networks with higher complexity, both in terms of size and dispersion. These results have significant implications for enhancing community detection in fields such as social and biological network analysis.
\end{abstract}


\end{frontmatter}



\section{Introduction}

Detecting community structures within complex networks, often referred to as community detection~\cite{fortunato2016community}, is a critical task in various fields, including social network analysis~\cite{maivizhi2016survey, huang2022information}, life sciences~\cite{atay2017community}, and computational systems~\cite{sangaiah2023explainable}. Identifying tightly interconnected nodes, common structure clusters, and information flows is essential for understanding the organizational and functional dynamics of network systems. Algorithms for community detection leverage the modularity of real-world networks, where intra-community connections are stronger than inter-community links~\cite{Fortunato2010}.

The motivation for this work arises from the complexity of real-world networks, which are often analyzed using a variety of community detection algorithms. Different algorithms can produce varied community structures, making it challenging to determine which divisions best represent the underlying network. This variability introduces uncertainty, particularly in regions where communities are not well defined or where boundaries between communities are unclear. Thus, there is a growing need for methods that can integrate these different perspectives into a unified solution, reducing uncertainty and providing a clearer understanding of the network structure.

In this context, we propose a community detection framework called Rough Clustering-based Consensus Community Detection (RC-CCD), which integrates rough set theory with consensus clustering. The use of rough set theory allows for the identification of "core" regions, where nodes belong to communities with high certainty, and "uncertain" regions, where nodes lie near the boundaries between communities. By aggregating the outputs of multiple community detection algorithms, the consensus clustering component enhances the accuracy of the identified communities, while rough set theory provides a framework to manage and interpret the uncertainty inherent in the data.

Existing approaches, such as matrix-based methods proposed by Lancichinetti et al.~\cite{lancichinetti2012consensus}, typically focus on producing a single partition of the network or do not adequately address the uncertainty associated with different community detection results. The RC-CCD framework addresses this gap by providing a consensus that reflects both the stability of core community members and the uncertainty at the boundaries between communities, resulting in a more comprehensive understanding of the network structure. This comprehensive understanding is particularly useful for analyzing overlapping communities, as it allows for better distinction between stable core regions and areas of uncertainty where nodes might belong to multiple communities.

The rest of the paper is structured as follows: Section~\ref{sec:related} reviews related work on community detection algorithms. Section~\ref{sec:RC-CCD} explains the RC-CCD methodology in detail. Sections~\ref{sec:experimentalSetup} and ~\ref{sec:results} present experimental results and compare the performance of RC-CCD against other community detection algorithms. These results are discussed in Section~\ref{sec:discussion}. Section~\ref{sec:conclusion} concludes the paper and suggests directions for future research.

\section{Related Work}\label{sec:related}
Existing community detection algorithms often fall short in accurately detecting said communities due to the unique characteristics of each network. This variation underlines the need for more sophisticated methods to refine detection accuracy. Addressing these challenges, consensus clustering, as highlighted by Lancichinetti et al. \cite{lancichinetti2012consensus} and Jeub et al. \cite{jeub2018multiresolution}, has marked a significant advancement, merging multiple clustering results into a unified representation and reducing algorithmic randomness and biases. 

One noteworthy example is the dual-level clustering ensemble algorithm \cite{duallevel2023}, which incorporates a comprehensive approach using three consensus strategies to generate a highly consistent outcome. Each type addresses the information from base clustering members differently, contributing uniquely to the production of the ensemble outcome.

Another example of the application of ensemble clustering in the discovery of cancer subtypes is the work by Parea et al. \cite{parea2023}, where a multi-view hierarchical ensemble clustering approach showed excellent performance in stratifying patients into sub-groups. The groups successfully mapped similar molecular characteristics across several types of cancer, outperforming current state-of-the-art methods in six out of seven cancer types. More innovative methods continue to emerge, aiming to refine clustering performance through ensemble and consensus strategies. Ji et al. \cite{Ji2022} introduced a clustering ensemble algorithm optimizing the accuracy of equivalence granularity to improve clustering quality by minimizing input data size and enhancing the diversity and accuracy of the base groupings. 

The exploration of consensus clustering in complex networks has significantly advanced our understanding of network structures. Lancichinetti and Fortunato \cite{lancichinetti2012consensus} were pioneers in demonstrating the efficacy of consensus clustering for enhancing the stability and accuracy of community detection in complex networks. Their methodology set a foundation for subsequent research by addressing the limitations inherent in single-resolution community detection methods. Inspired by this foundational work, Jeub et al. \cite{jeub2018multiresolution} introduced the concept of multiresolution consensus clustering. This approach not only acknowledges the multi-scale nature of community structures within networks, but also provides a framework for identifying these structures across different resolutions. Their methodology emphasizes the benefits of hierarchical consensus clustering on networks that exhibit complex, layered community structures. Further advancements in the field were made by Tandon et al. \cite{tandon2019fast}, who developed a fast consensus clustering technique that significantly reduces the computational demand of the consensus clustering process. This innovation enabled the application of consensus clustering to much larger networks than was previously feasible, marking a significant step forward in network analysis capabilities.

These contributions collectively highlight the evolving landscape of consensus clustering research, demonstrating its critical role in uncovering the nuanced community dynamics of complex networks. 

Building upon these significant contributions, our work overcomes prior limitations by integrating rough set theory with consensus clustering in the context of community detection in networks. By incorporating the rough set framework, our method addresses the challenge of managing uncertainty in community boundaries and provides a more refined representation of network structures. This integration not only differentiates our approach from matrix-based methods like those employed by Lancichinetti et al. \cite{lancichinetti2012consensus}, but also introduces a mechanism for balancing the certainty of core community members with the uncertainty at the boundaries.

The RC-CCD framework addresses this gap by delivering a consensus that captures both stable regions within the network and areas of uncertainty, resulting in a more comprehensive and nuanced understanding of the community structure. This allows for improved accuracy in detecting network communities, offering a deeper exploration of the dynamic interactions within complex networks.
\\

\section{Rough Clustering-Based Consensus Community Detection}\label{sec:RC-CCD}
\subsection{Networks as graphs}
A network is represented as an undirected graph \( G = (V, E) \), where \( V \) denotes the set of nodes with \( |V| = n \), and \( E \) the set of edges connecting these nodes with \( |E| = m \). The graph \( G \) is characterized by an adjacency matrix \( A \), with elements \( A_{ij} \) representing the presence (1) or absence (0) of an edge \( (v_i, v_j) \) between nodes \( i \) and \( j \). The degree of a node \( v_{i} \), denoted by \( d_{v_{i}} \), represents the number of edges connected to node \( v_{i} \).

Communities or clusters within a network are identified as subgraphs with high internal edge density while their external edge density is low. Formally, a community structure is a division \( \mathbb{P} = \{C_{1}, C_{2}, \ldots, C_{k}\} \) of the network into \( k \) subgraphs, where \( V = \bigcup_{i=1}^{k}C_{i} \).

\begin{definition}[Thresholded similarity graph]
\label{similarityGraph}
Given a set of nodes \( V = \{v_1, v_2, \ldots, v_n\} \) of a network \( G \), and a threshold \( \beta \), the thresholded similarity graph \( G_{\beta} \) is an undirected graph where an edge \( (v_i, v_j) \) exists if the similarity between nodes \( v_i \) and \( v_j \) is at least \( \beta \).
\end{definition}

\begin{definition}[Subgraph]
A graph \( G_1 \) is a subgraph of \( G_2 \), denoted as \( G_1 \subseteq G_2 \), if every node and edge in \( G_1 \) is also in \( G_2 \).
\end{definition}

\begin{definition}[Induced Subgraph]
For a graph \( G \) and a node subset \( V^{'} \subseteq V \), the induced subgraph \( G[V^{'}] \) is composed of \( V^{'} \) and all edges in \( G \) that connect pairs of nodes in \( V^{'} \).
\end{definition}

\begin{definition}[$\beta$-Connected Component]
\label{def:similarityGraph}
In a thresholded similarity graph \( G_{\beta} \), a subgraph \( G' \) is a \( \beta \)-connected component if every pair of distinct nodes in \( G' \) is connected directly or indirectly, and there is no larger subgraph containing \( G' \) that also satisfies this condition.
\end{definition}

\subsection{Rough Clustering: Extending Rough Set Theory for Clustering}
\label{roughClustering}

Rough Set Theory (RST) is based on the concepts of lower and upper approximations, which classify objects as either certainly or possibly belonging to a set \cite{yao2004granular}. This approach helps manage uncertainty in data by defining boundary regions between these approximations.

Lingras et al. \cite{lingras2004interval} extended RST with the rough \( k \)-means algorithm, where each cluster is represented by rough sets. Lower approximations contain objects that clearly belong to a cluster, while upper approximations capture objects that might belong to multiple clusters, allowing for overlaps. Key properties of this approach include:
\begin{itemize}
    \item Objects in the lower approximation belong to only one cluster.
    \item Objects in the lower approximation are also part of the upper approximation.
    \item Objects not in any lower approximation may belong to multiple upper approximations.
\end{itemize}

In our case, the term ``clusters'' refers to communities in the network. This approach is grounded in the broader framework of Granular Computing (GrC) \cite{yao2004granular}, which focuses on forming and processing granules—groups of similar objects. In this context, the rough \( k \)-means algorithm is used to refine these granules, enabling the identification of overlapping clusters and managing uncertainty by determining which objects clearly belong to a cluster and which may be shared between multiple clusters.

\subsection{RC-CCD Algorithm}
\label{sec:proposal}
Our proposed framework is based on the principles of Granular Computing \cite{yao2004granular}, applying the concepts of Rough Clustering, with a focus on the rough \( k \)-means algorithm \cite{lingras2004interval}, as described in the previous section. This method allows us to analyze network structures in terms of communities in a flexible and detailed manner.

The RC-CCD algorithm begins by taking as input a set of community partitions derived from either different community detection algorithms or multiple runs of a single algorithm with varying parameters. It also accommodates Pareto-optimal sets from multi-objective algorithms \cite{grass2020overlapping}, showcasing its adaptability to diverse community detection scenarios.

In this study, we employed algorithms such as Label Propagation (LPA) \cite{raghavan2007near}, Greedy Modularity \cite{clauset2004finding}, Infomap \cite{rosvall2008maps}, and Louvain \cite{blondel2008fast}, selected for their computational efficiency and scalability to large networks. These algorithms provide the community partitions that form the foundation of the framework.

It is important to note that the computational complexity of RC-CCD is independent of the algorithms used to generate the input partitions. While these algorithms supply the necessary initial partitions, their computational cost is not included in the analysis of RC-CCD’s performance.

The method begins by grouping similar nodes in the network \( G \) to form granules based on their co-location in communities across partitions. This process generates a partition of the node set \( V \), establishing equivalence classes that uncover node relationships, which are essential for defining community structures.


In a second phase, these granules are treated as rough sets, allowing the identification and representation of overlapping communities. This phase refines the granules using the rough \( k \)-means algorithm, ensuring a precise and realistic depiction of network communities. The GrC framework is leveraged to define the ``granulation criterion'' \cite{yao2004granular}, which in our case is based on the frequency with which nodes appear together in the same communities across different partitions. The rough \( k \)-means algorithm then refines these granules, effectively capturing the network structures, highlighting both the stable core regions and the more uncertain overlapping areas.


\subsubsection{First step: Identifying granules of indiscernible nodes}

In this initial phase, our method identifies granules of inseparable nodes by constructing a thresholded similarity graph \( G_{\beta} \). This graph captures node relationships based on their co-occurrence in communities across the input partitions. A similarity function measures the frequency with which nodes appear together, providing a basis for the strength of their connection.

The connected components of \( G_{\beta} \) reveal clusters of nodes that are closely related according to this similarity measure. By taking these clusters and their induced subgraphs in the original graph \( G \), we obtain the granules of indiscernible nodes.

\paragraph{Nodes similarity measure}
Let \(\mathbb{NP} = \{\mathbb{P}_{1}, \mathbb{P}_{2}, \dots, \mathbb{P}_{p}\}\) be a set of network partitions of $G$, comprising partitions from various runs of algorithms. Notably, each algorithm is expected to produce distinct partitions based on a variety of mathematical foundations. Here, \(\mathbb{P}_r\) represents an individual partition within \(\mathbb{NP}\) for $1 \leq r \leq p$, where $p$ equals the total number of runs across all algorithms.

We define the similarity function \(S_{_{\mathbb{NP}}}(v_i, v_j)\) to measure the closeness of nodes $v_i$ and $v_j$ based on their co-occurrence across $\mathbb{NP}$. This function quantifies the degree of association between two nodes within the context of the given network partitions. 

Let \(\mathbb{CS}_{v_i}\) be the set of communities where node \(v_i\) is included. We define \(mc(v_i, v_j)\) as the number of communities within the partition \(\mathbb{P}_r\) in which both nodes \(v_i\) and \(v_j\) are jointly present. It counts how many communities include both nodes as members, indicating their level of co-membership in shared communities.

This definition is a formalization of the elements \(i, j\) of the consensus matrix \(D\) from the work of Lancichinetti et al \cite{lancichinetti2012consensus}. The similarity function $S_{_{\mathbb{NP}}}(v_i,v_j)$ is then defined as:
\begin{equation}
\label{eq:simVertexNP}
S_{_{\mathbb{NP}}}(v_i,v_j) = \frac{\sum_{\mathbb{P} \in \mathbb{NP}} match(v_i,v_j)}{\vert \mathbb{NP} \vert},
\end{equation}
where $match(v_i,v_j)$ is calculated by:
\begin{equation}
match(v_i,v_j) = \frac{mc(v_i,v_j)}{\vert \mathbb{CS}_{v_i} \vert \cdot \vert \mathbb{CS}_{v_j} \vert}.
\end{equation}

\paragraph{Indiscernible nodes granules}
The thresholded similarity graph \( G_{\beta} \) is constructed using the similarity measure \( S_{_{\mathbb{NP}}}(v_i, v_j) \) and a user-defined threshold \( \beta \) within \([0,1]\). In \( G_{\beta} \), edges indicate how frequently nodes \( v_i \) and \( v_j \) co-occur in the same communities across the network partitions \( \mathbb{NP} \). The connected components of \( G_{\beta} \), denoted as \( G_{r}^{'} = \{ G_{r_1}^{'}, G_{r_2}^{'}, \ldots, G_{r_q}^{'} \} \), identify groups of nodes that are closely related based on this similarity.

By mapping these connected components back to their corresponding induced subgraphs in the original graph \( G \), we obtain \( G_r = \{ G_{r_1}, G_{r_2}, \ldots, G_{r_q} \} \), representing an effective partitioning of the node set \( V \). Each subgraph in \( G_r \) corresponds to a group of inseparable nodes that exhibit consistent clustering patterns from the input partitions \( \mathbb{NP} \). 

This approach simplifies the network structure by grouping similar nodes into granules, providing a foundation for the next phase where computations are performed using rough clustering concepts.

\subsubsection{Second step: build the final covering of $G$}
The second phase applies the rough \( k \)-means algorithm \cite{lingras2004interval} to refine network coverage. This step is specifically designed to adjust the regions of low uncertainty and those of higher uncertainty, with the latter often representing the overlapping areas between the resulting communities. We select \( k \) community prototypes from \( G_{r} \) (where \( 1 \leq k \leq q \)) based on node count, integrating the remaining granules to effectively depict the network's complexity. This phase leverages lower and upper approximations to provide a comprehensive representation of community structures.

\paragraph{Selection of $k$ prototypes}
Our method selects \( k \) core prototypes for consensus communities based on input from the algorithms in Section~\ref{sec:proposal} to form a collection \(\mathbb{NP}\), each yielding partitions \( \mathbb{P}_r \) with \( k_{_{r}} \) communities. These algorithms often identify larger communities first, influencing the selection process. We choose \( k \) based on the number of communities covering the most network nodes, ensuring that these prototypes represent significant network areas.

A cumulative frequency histogram is created from these partitions, merging bars representing the same \( k_{_{r}} \)-th community. The histograms are ordered by frequency, and \( k \) is determined by the count of the top histograms covering $90\%$ of the total frequency. The top \( k \) communities are then selected as prototypes for our consensus clustering.

\paragraph{Assign the remaining granules}
After the initial \( k \) prototype selection, our method efficiently assigns the remaining granules to these cores, using two defined similarity functions for precise community delineation between any two granules \( G_{r_i}, G_{r_j} \in G_r \). At this stage, \( G_{r_i} \) represents the \( i \)-th prototype (where \( 1 \leq i \leq k \)), and \( G_{r_j} \) denotes the \( j \)-th remaining granule to be assigned (where \( k+1 \leq j \leq q \)).

The first similarity function, \( S_{G_r}(G_{r_i}, G_{r_j}) \), measures granule similarity by assessing node co-location in the same community across partitions in \( \mathbb{NP} \):
\begin{equation}
S_{G_r}(G_{r_i}, G_{r_j}) = \sum_{v_i \in G_{r_i}} \sum_{v_j \in G_{r_j}} S_{\mathbb{NP}}(v_i, v_j)
\label{eq:SimGranulesmatrix}
\end{equation}
where \( S_{\mathbb{NP}}(v_i, v_j) \) is the nodes similarity measure defined in Equation \ref{eq:simVertexNP}.

The second function, \( S_{E_r}(G_{r_i}, G_{r_j}) \), quantifies the connections between nodes within granules:
\begin{equation}
S_{E_r}(G_{r_i}, G_{r_j}) = \sum_{v_i \in G_{r_i}} \sum_{v_j \in G_{r_j}} A_{v_i, v_j}
\label{eq:SimGranulesconnections}
\end{equation}
where \( A \) is the adjacency matrix of the network, and \( A_{v_i, v_j} \) indicates the presence (1) or absence (0) of an edge between nodes \( v_i \) and \( v_j \).

For each prototype \( G_{r_i} \), we compute \( S_{G_r}(G_{r_i}, G_{r_j}) \) for all \( j \)-th granules and normalize these values within \([0,1]\) using the maximum value of \( S_{G_r} \), denoted as \( S_{G_r}^{\text{max}} \). A similar normalization is applied to \( S_{E_r} \), yielding normalized values with respect to \( S_{E_r}^{\text{max}} \). We then derive a composite similarity, \( CS_{G_{r_j}} \), as the average of these normalized values for each granule \( G_{r_j} \):
\begin{equation}
CS_{G_{r_j}} = \frac{1}{2} \left( \frac{S_{G_r}(G_{r_i}, G_{r_j})}{S_{G_r}^{\text{max}}} + \frac{S_{E_r}(G_{r_i}, G_{r_j})}{S_{E_r}^{\text{max}}} \right)
\end{equation}

Each granule \( G_{r_j} \) is assigned to community approximations based on its score $CS_{_{G_{rj}}}$ and a threshold \(\gamma\) within \([0,1]\). Granules exceeding \(\gamma\) in similarity to a prototype \( G_{r_i} \) join the lower approximation, indicating strong alignment, while others enter the upper approximation for weaker associations. This method effectively delineates community structures, capturing both clear and ambiguous node relationships in the network.

\begin{algorithm}[htpb]
\scriptsize 
\caption{RC-CCD pseudo-code.}
\label{alg:RC-CCD}
\SetAlgoLined
\KwIn{Network $\red$, Set of Network Partitions $\mathbb{NP}$}
\KwOut{Network Coverage $CV = \{CV_1, CV_2, \ldots, CV_k\}$}
\textbf{Step 1: Build Granules of Indiscernible nodes}\;
\For{each pair of nodes $v_i, v_j \in V$}{
    Compute similarity $S_{_{\mathbb{NP}}}(v_i, v_j)$ using $\mathbb{NP}$\ and $\beta$ parameter;
}
Construct thresholded similarity graph $G_{\beta} = (V, E_{\beta})$\;
Identify $\beta$-connected components in $G_{\beta} = (V, E_{\beta})$\;
Build $G_r = \{G_{r_1}, G_{r_2}, \ldots, G_{r_q}\}$, subgraphs induced in $G$ from $G_{r}^{'}$\;
\textbf{Step 2: Build Final Network Coverage}\;
Sort $G_r$ in descending order by node count\;
Select top $k$ granules $G_{r_i} \in G_{r}$ as prototypes for communities $CV_i$\;
\For{$i \leftarrow 1$ \KwTo $k$}{
    Assign $G_{r_i}$ to $CV_i$\;
}

\For{$j \leftarrow k+1$ \KwTo $q$}{
    $T \leftarrow \{ \}$\;
    \For{$i \leftarrow 1$ \KwTo $k$}{
        \If{$CS_{G_{r_j}} (G_{rj}, G_{ri})\geq \gamma$}{
            Add $G_{ri}$ to $T$\;
        }   
    }
    \If{$|T| > 1$}{
        $\forall G_{ri} \in T$ take the community $CV_i$ associated\;
        Add $G_{rj}$ to $\overline{CV_i}$\;
    }
    \Else{
        Take the community $CV_{i}$ associated with $G_{ri} \in T$\;
        Add $G_{rj}$ to $\underline{CV_{i}}$ and $\overline{CV_{i}}$\;
    }
}
\Return $CV$\;
\normalsize 
\end{algorithm}

\paragraph{RC-CCD Algorithm: Pseudocode Overview and Step-by-Step Process}
The proposed framework, detailed in Algorithm \ref{alg:RC-CCD} (code available at \footnote{\url{https://github.com/Leandroglez39/RoughSetsConsensusClustering}}), begins with a set of network partitions generated by previously executed algorithms. Although there is no strict criterion for selecting these algorithms, using a diverse set provides multiple perspectives on community structures, which enriches the consensus process and enhances robustness and accuracy.

In the first phase, the method calculates a similarity function between nodes. This function, along with a user-defined parameter \( \beta \), is used to construct the \(\beta\)-similarity graph \( G_{\beta} \). The connected components of this graph identify clusters of inseparable nodes, forming granules within the original graph \( G \).

The second phase focuses on building the final network covering. It begins by ordering the granules of inseparable nodes, \( G_{r} \), based on their size. From this ordered set, \( k \) granules are selected as prototypes for the communities. The remaining granules in \( G_{r} \) are then assigned to either the lower or upper approximations of these prototypes, based on their similarity scores \( CS_{_{G_{rj}}} \) and a parameter \( \gamma \).

\begin{figure}[H]
    \centering
    \includegraphics[width=0.96\linewidth]{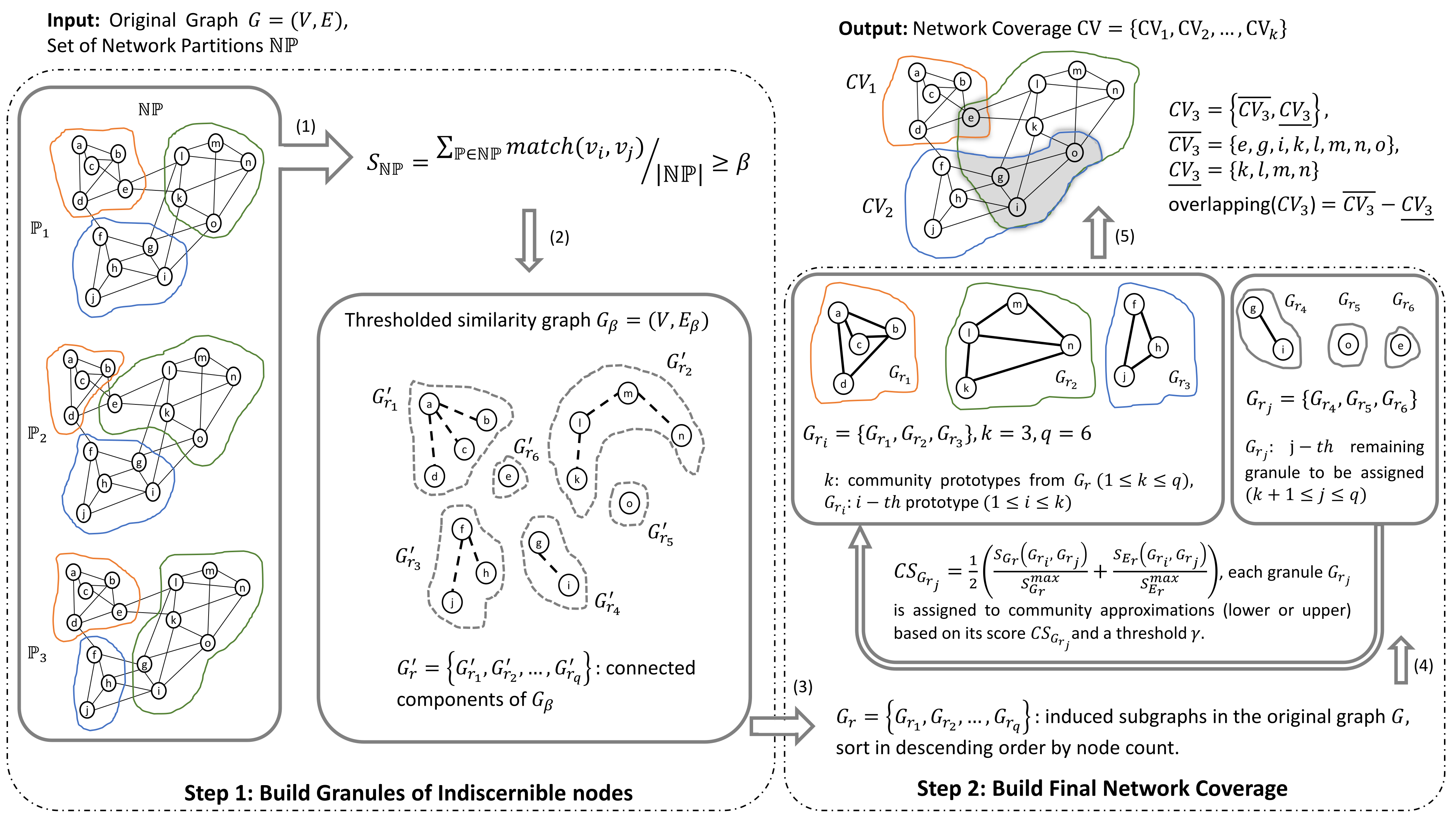}
    \caption{Summary of the RC-CCD process: (1) Compute the similarity $S_{_{\mathbb{NP}}}$ between vertex pairs based on shared communities across partitions $\mathbb{NP}$; (2) Construct the \(\beta\)-similarity graph and identify its connected components; (3) Identify induced subgraphs in the original graph as sets of inseparable nodes; (4) Select the largest \(k\) granules as community prototypes, and assign remaining granules to lower or upper approximations using the similarity function \( CS_{_{G_{rj}}} \) and \(\gamma\) threshold; (5) Output communities with core and boundary members.}
    \label{fig:procedure}
\end{figure}

RC-CCD’s efficiency is primarily determined by the construction of the similarity graph and the refinement of granules. The method innovatively adapts the \(\beta\)-thresholded graph and consensus matrix approach from Lancichinetti et al. \cite{lancichinetti2012consensus}, improving accuracy by analyzing connection strengths both within and between communities. A visual summary of this procedure is provided in Figure \ref{fig:procedure}.

\section{Experimental Setup}\label{sec:experimentalSetup}
\subsection{Base Algorithms \& Metrics}\label{sec:exp_setup}
In this work, we employed four state-of-the-art community detection algorithms as both inputs and benchmarks, which we will refer to as base algorithms, to test our RC-CCD method \cite{blondel2008fast,clauset2004finding,raghavan2007near,rosvall2008maps}.
A brief description of each algorithm follows.

\begin{itemize}
    \item \textbf{Louvain} is a hierarchical clustering algorithm with a complexity of \(\mathcal{O}(n \cdot \log n)\). Effective at optimizing network modularity and identifying diverse community sizes, it may struggle with small communities in the presence of larger ones. It converges when modularity cannot be further improved~\cite{blondel2008fast}.
    \item \textbf{Greedy Modularity} is an agglomerative algorithm similarly focused on optimizing modularity, with a computational complexity of \(\mathcal{O}(m \cdot \log^2 n)\). While effective, its computational demand makes it less suitable for large networks and it faces challenges in detecting smaller communities. Convergence occurs when no community fusion increases modularity~\cite{clauset2004finding}.
    \item \textbf{Label Propagation (LPA)} is a non-deterministic algorithm with a linear complexity of \(\mathcal{O}(m)\). LPA tends to identify balanced-sized communities but may overlook smaller ones in dense networks. It stabilizes when a node's label matches the majority of its neighbors'~\cite{raghavan2007near}.
    \item \textbf{Infomap} uses information theory with efficient time performance, capable of detecting various community sizes, including nested and overlapping ones. It might yield very small communities and stabilizes when the description length of the random walk is minimized~\cite{rosvall2008maps}.
\end{itemize}

Two well-known graph metrics were used to assess performance and compare the output communities across runs of each algorithm.

\textbf{Normalized Mutual Information (NMI)} is a commonly used metric in the field of network analysis and community detection \cite{lancichinetti2009detecting}. It serves to evaluate the performance of community detection algorithms by measuring the similarity between two different community assignments. Specifically, it compares a ground truth community structure with the outcome community structure from an algorithm, taking into account both the homogeneity and completeness of the assignments.

The NMI value lies in the range of 0 to 1. A value of 1 signifies a perfect match, while a value of 0 indicates no similarity between the two community assignments. A higher NMI value thus implies a stronger agreement between the community structures under comparison, serving as a quantitative measure for evaluation and comparison of different methods.

\textbf{Participation Coefficient (PC)} is another well-used metric in the community detection field \cite{guimera2005functional}. It quantifies the diversity of a node's connections across different communities in a network. The PC for a node \(v_{i}\) is given by:

\begin{equation}
P(v_{i}) = 1 - \sum_{s=1}^{k} \left( \frac{d_{v_{i},s}}{d_{v_{i}}} \right)^2
\end{equation}

Here \(d_{v_{i},s}\) represents the number of links from node \(v_{i}\) to nodes in community \(s\), \(d_{v_{i}}\) is the node's total degree, and \(k\) is the total number of communities. To extend the applicability of the PC to networks with overlapping communities, the total degree \(d_{v_{i}}\) of a node is modified as follows:

\begin{equation}
d_{v_{i}} = \sum_{c \in C} h_{v_{i},c}
\end{equation}

where \(C\) represents the set of communities and \( h_{v_{i},c}\) is the count of node \(i\)'s neighbors that also belong to community \(c\).

\subsection{Synthetic Networks}
Our study uses the $LFR$ framework to generate test networks of various power-law distributions in node degree and community size, reflecting real-world network characteristics \cite{LancichinettiFortunato2009}. The $LFR$ framework can accurately replicate complex network structures while providing the truth communities for comparison. Key parameters when generating the networks include node degrees and community sizes, governed by \(\tau_{1}\) and \(\tau_{2}\), as well as network size \(N\), average degree \(k\), maximum degree \(k_{\text{max}}\), and community size range limits \(c_{\text{min}}\) and \(c_{\text{max}}\). A key feature of $LFR$ is the mixing parameter \(\mu\) that controls the fraction of inter-community edges, with lower values yielding clearer clusters and higher values resulting in mixed communities. For those networks with overlapping communities, two other parameters are also key: \(O_{n}\) as the overlap rate, and \(O_{m}\) as the node participation in multiple communities.

In our experiments, we configured two distinct sets of parameters to assess the influence of network size on community detection. We varied \( \mu \) from 0.1 to 0.6 in increments of 0.05, generating 11 distinct networks for each configuration, labeled from \( net1 \) to \( net11 \). These labels will be consistently used in subsequent tables and figures to denote the specific networks analyzed. The full configurations for the two types of networks were as follows:

\begin{itemize}
    \item Small network configuration: \(n = 1000\), \(\tau_{1} = 2\), \(\tau_{2} = 1\), \(k = 15\), \(k_{\text{max}} = 50\), \(c_{\text{min}} = 20\), \(c_{\text{max}} = 50\), \(O_{n} = 100\), \(O_{m} = 2\)
    \item Large network configuration: \(n = 20000\), \(\tau_{1} = 2\), \(\tau_{2} = 1\), \(k = 15\), \(k_{\text{max}} = 50\), \(c_{\text{min}} = 40\), \(c_{\text{max}} = 100\), \(O_{n} = 2000\), \(O_{m} = 2\)
\end{itemize}

\section{Results}
\label{sec:results}
This section provides a comprehensive evaluation and comparison of the RC-CCD algorithm and the base algorithms from Section~\ref{sec:exp_setup}. The evaluation of RC-CCD included NMI for accuracy, analysis of variability in community detection or outcome stability, and a modified participation coefficient for overlapping communities. All evaluations were conducted on $LFR$-generated synthetic networks.

The parameters \( \beta \) and \( \gamma \) in RC-CCD serve as mechanisms to adjust the level of certainty in identifying the core (lower approximation) and boundary (upper approximation) regions of the communities. These parameters provide adaptability depending on the domain of application. In this study, we set \( \beta = 0.75 \) and \( \gamma = 0.8 \), values recommended in previous research \cite{lingras2012applying, lingras2004interval, mitra2004evolutionary}. Additionally, we performed a detailed analysis on the effect of the \( \gamma \) parameter, which is critical for detecting community boundaries at different scales. This allowed us to fine-tune the identification of both well-defined community cores and more ambiguous boundary regions.

\subsection{NMI quality measure}
\begin{figure}[h]
\begin{center}
\subfigure[Small Networks ($n=1000$).]{\includegraphics[trim = 1mm 1mm 1mm 8mm, clip, width=65mm]{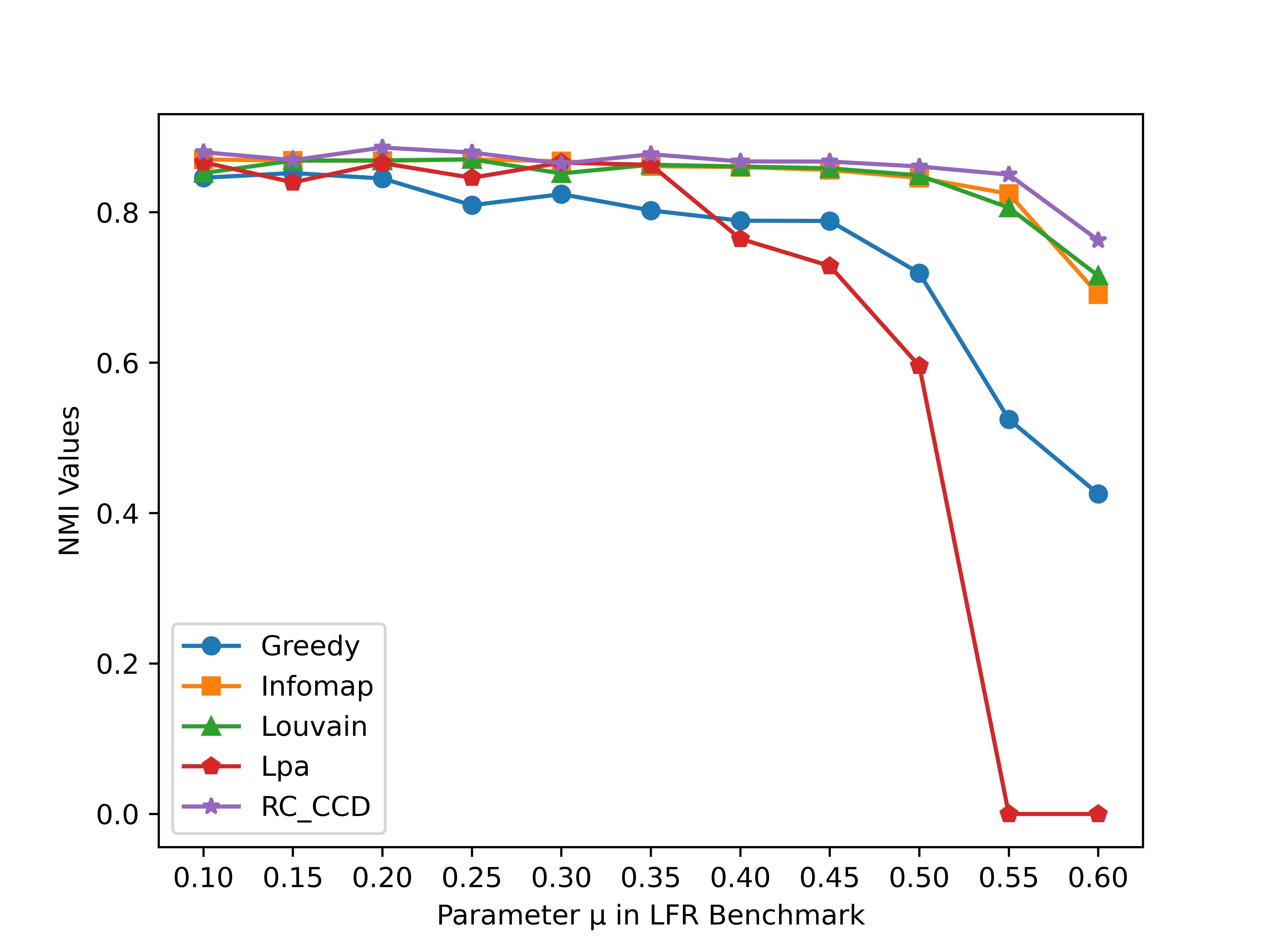}\label{dataset:small}}
\subfigure[Large Networks ($n=20000$).]{\includegraphics[trim = 1mm 1mm 1mm 8mm, clip, width=65mm]{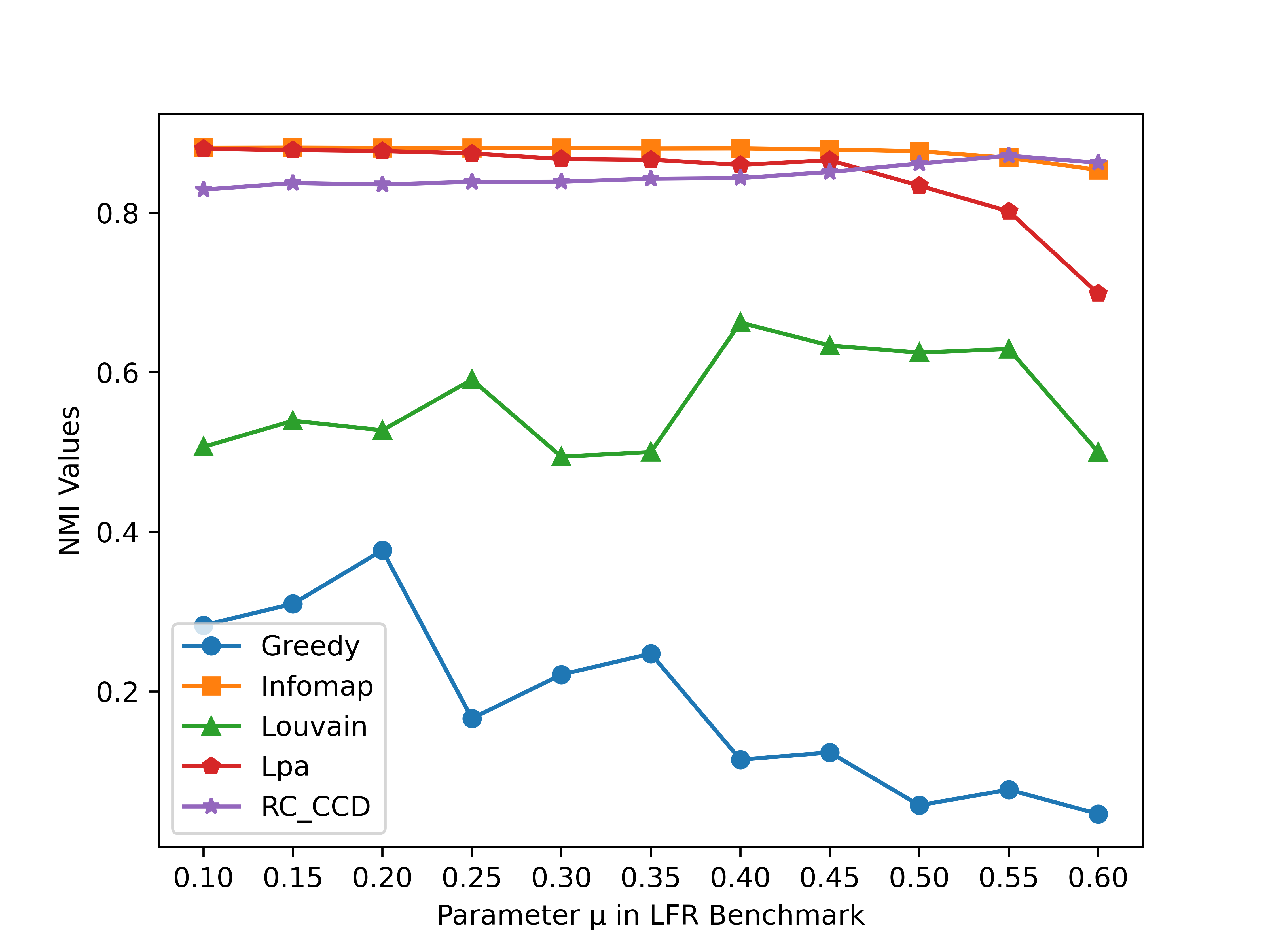}\label{dataset:large}}
\caption{Comparison of the base algorithms and RC-CCD on $LFR$ benchmark.}\label{img:nets_NMI_score}
\end{center}
\end{figure}

We firstly assessed and compared the accuracy of the RC-CCD and the base algorithms' outcomes based on the NMI metric (see \ref{sec:exp_setup}), including small (\(n = 1000\)) and large networks (\(n = 20000\)). A visual summary of the NMI scores across algorithms is shown in Figure~\ref{img:nets_NMI_score}. Extended NMI results are included Tables~\ref{tab:NMI_Values_SmallNets} and \ref{tab:NMI_Values_LargeNets} in the \textit{Supplementary Material}.

For small networks, RC-CCD consistently achieves high NMI values, often above 0.85, indicating strong alignment with ground-truth communities. While base algorithms Infomap and Louvain show robust performance, LPA notably scores an NMI of zero in \(net10\) ($\mu = 0.55$) and \(net11\) ($\mu = 0.6$), suggesting a complete mismatch with the actual community structure. The Greedy algorithm generally under performs across all runs.

For large networks, RC-CCD consistently achieves results that are equal to or better than the other evaluated algorithms. For example, in \(net1\), the NMI score of 0.829 for RC-CCD is close to Infomap's 0.882 and significantly higher than Greedy's 0.283. Notably, in more complex networks such as \(net10\) and \(net11\), RC-CCD achieves slightly higher values than Infomap, which consistently maintains high scores across networks. In contrast, the Greedy algorithm continues to perform poorly, maintaining low values throughout.

In summary, RC-CCD demonstrates reliable and accurate community detection in diverse networks, as evidenced by consistent high NMI scores (see Tables~\ref{tab:NMI_Values_SmallNets} and \ref{tab:NMI_Values_LargeNets}). Its robustness and adaptability are further validated through comprehensive comparisons with base algorithms, as shown in Figure~\ref{img:nets_NMI_score}.

\subsection{Stability}
Stability, essential for measuring algorithm consistency across multiple runs \cite{kwak2011consistent}, was quantitatively assessed using the NMI index to track community assignment consistency by all the algorithms.

Small and large network configurations were tested through 10, 50, and 100 runs, each replicated \( r = 20 \) times, to confirm result consistency. Mean NMI scores from these repetitions served as indicators of stability, with the modularity metric used to identify distinct community structures from multiple algorithm outputs.

Summarized in Table~\ref{tab:tableStability1} and ~\ref{tab:tableStability2}, the evaluation on complex networks (\( \mu = 0.6 \)) showed that in small configurations, LPA was unstable with zero NMI (Table~\ref{tab:tableStability1}). Greedy averaged \( 0.43 \pm 0.05 \) in NMI, while Louvain and Infomap showed moderate stability with \( 0.71 \pm 0.03 \) and \( 0.67 \pm 0.04 \) respectively. RC-CCD demonstrated superior stability with an average NMI of \( 0.76 \pm 0.01 \).

\begin{table}[htbp]
\centering
\small 
\begin{tabular}{|l|ccc|}
\cline{2-4}
\multicolumn{1}{c|}{} &  \multicolumn{3}{c|}{$\mu = 0.6$} \\
\hline
Alg & 10 & 50 & 100 \\
\hline
LPA & 0.00 $\pm$ 0.00 & 0.00 $\pm$ 0.00 & 0.00 $\pm$ 0.00 \\
Greedy & 0.43 $\pm$ 0.05 & 0.41 $\pm$ 0.05 & 0.43 $\pm$ 0.04 \\  
Louvain & 0.71 $\pm$ 0.03 & 0.70 $\pm$ 0.03 & 0.70 $\pm$ 0.03 \\
Infomap & 0.67 $\pm$ 0.04 & 0.66 $\pm$ 0.06 & 0.68 $\pm$ 0.07 \\
RC-CCD & \textbf{0.76 $\pm$ 0.01} & \textbf{0.73 $\pm$ 0.01} & \textbf{0.76 $\pm$ 0.00} \\
\hline
\end{tabular}
\caption{Small Network Configuration: NMI quality and standard deviation comparison of RC-CCD and base algorithms for network complexity $\mu = 0.6$}
\label{tab:tableStability1}
\end{table}

In larger networks, RC-CCD maintained high stability with NMI scores between 0.86 and 0.87 (see Table~\ref{tab:tableStability2}). Infomap slightly decreased to around 0.85 NMI, while Louvain improved to 0.63-0.64 NMI. LPA showed significant improvement from its small network instability, achieving NMI values in the 0.69-0.71 range. Conversely, the performance for the Greedy alternative dropped to a low of 0.03 NMI.

\begin{table}[htbp]
    \centering
    \small 
    \begin{tabular}{|l|ccc|}
    \cline{2-4}
    \multicolumn{1}{c|}{} &  \multicolumn{3}{c|}{$\mu = 0.6$} \\
    \hline
    Alg & 10 & 50 & 100 \\
    \hline
    LPA & 0.71 $\pm$ 0.00 & 0.71 $\pm$ 0.00 & 0.69 $\pm$ 0.00 \\
    Greedy & 0.03 $\pm$ 0.00 & 0.03 $\pm$ 0.00 & 0.03 $\pm$ 0.00 \\
    Louvain & 0.63 $\pm$ 0.00 & 0.64 $\pm$ 0.00 & 0.62 $\pm$ 0.00 \\
    Infomap & 0.85 $\pm$ 0.00 & 0.85 $\pm$ 0.00 & 0.85 $\pm$ 0.00 \\
    RC-CCD & \textbf{0.86 $\pm$ 0.00} & \textbf{0.86 $\pm$ 0.00} & \textbf{0.87 $\pm$ 0.00} \\
    \hline
    \end{tabular}
    \caption{Large Network Configuration: NMI quality and standard deviation comparison of RC-CCD and base algorithms for network complexity $\mu = 0.6$}
    \label{tab:tableStability2}
\end{table}

Notably, LPA performance shifted from unstable ranges in small networks to moderate stability in larger ones, while RC-CCD and Infomap maintained high values across network sizes, showcasing their robustness. Detailed stability summaries are detailed in Supplementary Material Table~\ref{tab:summarynmismall} and Table ~\ref{tab:summarynmilarge}. An interesting observation across all algorithms was the absence of standard deviation in the NMI scores, indicating a highly consistent performance within each method across multiple runs. These stability results also highlighted the impact of network complexity (\( \mu \)) on the algorithms performance. Even with these varying network complexities, RC-CCD emerged as a highly reliable and adaptable choice.

\subsection{Cores Accuracy Evaluation}
Our research enhances classical community detection approaches by using a lower approximation to find the core elements of each community from the $k$ communities determined by multiple algorithm runs. The accuracy in finding these cores when compared to the ground-truth distribution (GT) for both small and large networks is presented in Table~\ref{tab:Unified_Coreaccuracy}.

\begin{table}[htbp]
    \centering
    \small 
    \begin{tabular}{|c|ccccccccccc|}
    \toprule
    Nets & net1 & net2 & net3 & net4 & net5 & net6 & net7 & net8 & net9 & net10 & net11 \\
    \midrule
    Small & 93.7 & 93.9 & 92.6 & 94.7 & 93.4 & 93.8 & 93.3 & 90.0 & 89.7 & 96.5 & 96.1 \\
    Large & 92.8 & 95.5 & 94.9 & 95.2 & 94.2 & 95.0 & 93.2 & 93.8 & 91.7 & 90.6 & 83.6 \\
    \bottomrule
    \end{tabular}
    \caption{Accuracy of RC-CCD's lower approximation when finding ground-truth communities in small and large networks}
    \label{tab:Unified_Coreaccuracy}
\end{table}

The results across small and large network configurations revealed that the proposed method for determining lower approximations consistently reached accuracy values over 90\%. The evaluation in Table~\ref{tab:Unified_Coreaccuracy} also showed that the method is consistent across varying degrees of network complexity.

The analysis of the number of communities \( k \) identified by the RC-CCD method is pivotal. In Table~\ref{tab:Community_Count_Comparison} within the Supplementary Material, we compare the counts of communities detected by RC-CCD to those of the GT. This comparison across 11 networks with varied complexities highlights the accuracy of our method in identifying community structures. The results show clear trends in how the complexity of the network influences the precision of the community detection by RC-CCD.

In small networks, our approach matched the GT for community number (\(k\)) across networks net1 to net9 (\(k=31\)), showcasing its strength in accurately identifying community structures in less connected environments. For larger networks, the method showed enhanced precision in complex settings, especially in \(net10\) and \(net11\) where it  closely approximated the GT. The RC-CCD method demonstrated outstanding performance on the network \(net11\), identifying \(k=366\) communities, compared to \(k=337\) communities detected by the GT. This result underscores the adaptability and efficiency of RC-CCD in handling networks of larger and more complex structures.

\subsection{Evaluation of Community Overlap}
Although our algorithm was not explicitly designed to identify overlapping communities, its mathematical formulation allows for it. This represents a significant advancement beyond previous studies \cite{lancichinetti2012consensus, jeub2018multiresolution}, as it identifies nodes that can belong to multiple communities. In our experiments, we configured the $LFR$ benchmark engine with a \(10\%\) overlapping rate to evaluate RC-CCD's performance in this task. Participation Coefficient (PC) was used to assess the degree of a node's involvement across different communities. We compared the nodes included in the overlapping communities identified by our method with those in the overlapping communities from the ground-truth. Higher PC values indicate that a node is active in multiple communities, and our results confirmed that RC-CCD effectively identifies these overlapping nodes.

Table \ref{tab:Unified_ParticipationCoefficient} presents the average PC degree for both ground truth and RC-CCD results across networks of varying complexities (\(\mu\)), for both small ($n=1000$) and large ($n=20000$) networks. These values progressively increased with network complexity, reflecting RC-CCD improved detection of overlapping communities as the network becomes more intricate.

\begin{table}[htbp]
\centering
\small 
\begin{tabular}{c c c c c c c c c}
\toprule
Net & \multicolumn{2}{c}{PC\_Mean\_GT} & \multicolumn{2}{c}{PC\_Mean\_RC} & \multicolumn{2}{c}{T\_Positive} & \multicolumn{2}{c}{F\_Positive} \\
\cmidrule(r){2-3} \cmidrule(r){4-5} \cmidrule(r){6-7} \cmidrule(r){8-9}
    & Small & Large & Small & Large & Small & Large & Small & Large \\
\midrule
net1 & 0.65 & 0.66 & 0.61 & 0.64 & 9  & 208 & 0  & 36   \\
net2 & 0.70 & 0.70 & 0.69 & 0.66 & 3 & 221 & 0 & 0  \\
net3 & 0.73 & 0.73 & 0.68 & 0.73 & 16 & 189 & 0  & 102  \\
net4 & 0.75 & 0.75 & 0.70 & 0.76 & 10  & 124 & 0  & 77  \\
net5 & 0.77 & 0.78 & 0.76 & 0.79 & 11 & 167 & 0  & 0  \\
net6 & 0.80 & 0.80 & 0.78 & 0.81 & 13 & 195 & 0 & 36  \\
net7 & 0.82 & 0.82 & 0.81 & 0.82 & 8 & 204 & 2 & 40 \\
net8 & 0.84 & 0.84 & 0.81 & 0.84 & 10 & 215 & 0 & 4 \\
net9  & 0.85 & 0.86 & 0.82 & 0.84 & 16 & 231 & 1  & 47  \\
net10 & 0.88 & 0.88 & 0.84 & 0.86 & 14 & 216 & 6 & 21  \\
net11 & 0.88 & 0.89 & 0.85 & 0.87 & 20 & 256 & 16 & 87 \\
\bottomrule
\end{tabular}
\caption{Comparative Analysis of mean Participation Coefficient values and True/False Positives for RC-CCD Against ground-truth communities in small and large networks.}
\label{tab:Unified_ParticipationCoefficient}
\end{table}

For smaller networks, the RC-CCD method's PC values were slightly lower (e.g., 0.61 for RC-CCD versus 0.65 for GT in net1) than in more complex networks (e.g., 0.85 for RC-CCD versus 0.88 for GT in net11). A similar trend was observed in larger networks, where initial PC values were marginally lower for RC-CCD but approach ground truth values as network complexity increased (e.g., from 0.64 for RC versus 0.66 for GT in net1, to 0.87 for RC-CCD versus 0.89 for GT in net11).

Additionally, we conducted a thorough comparison of our method's performance in detecting overlapping nodes against the ground truth, across different network sizes. This analysis, detailed in Table \ref{tab:Unified_ParticipationCoefficient}, included assessments of both true and false positive rates. In smaller networks like \(net1\), our method showed precision by identifying 9 out of 100 overlapping nodes with no false positives. Notably, for small networks, zero false positives were recorded in networks \(net2\), \(net3\), \(net4\), \(net5\), and \(net8\), demonstrating good accuracy across a spectrum of complexities. This result confirmed that the detected overlaps by RC-CCD were indeed accurate, even though it captured a modest number of true positives. Conversely, in larger networks like \(net11\), there was a significant increase in true positive detection --256 out of 2000 nodes-- along with an increase in false positives to 87.

\begin{figure*}[!h]
\begin{center}
\subfigure[Small Network ($n=1000$).]{\includegraphics[trim = 1mm 1mm 1mm 8mm, clip, width=65mm]{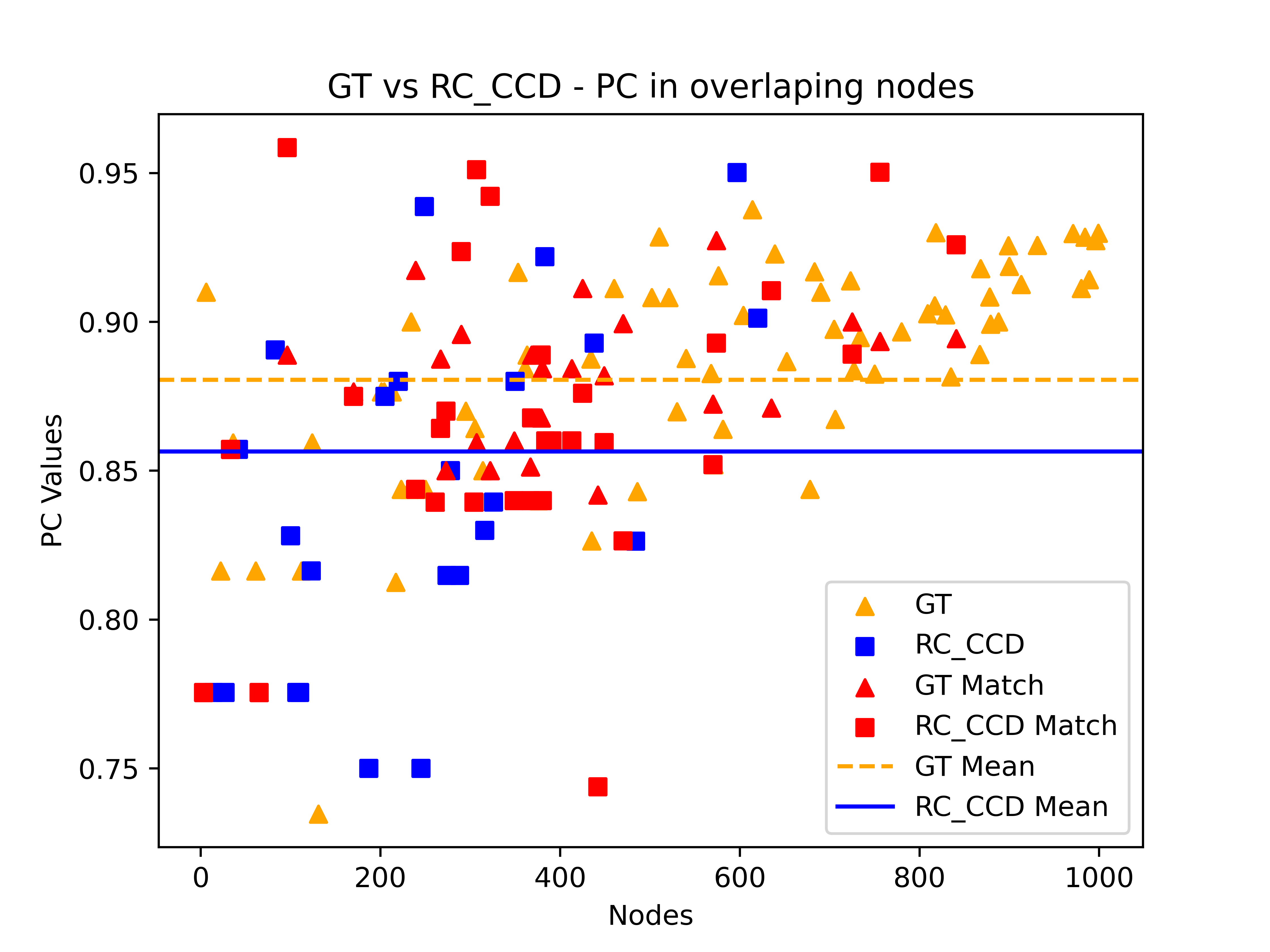}}
\subfigure[Large Network ($n=20000$).]{\includegraphics[trim = 1mm 1mm 1mm 8mm, clip, width=65mm]{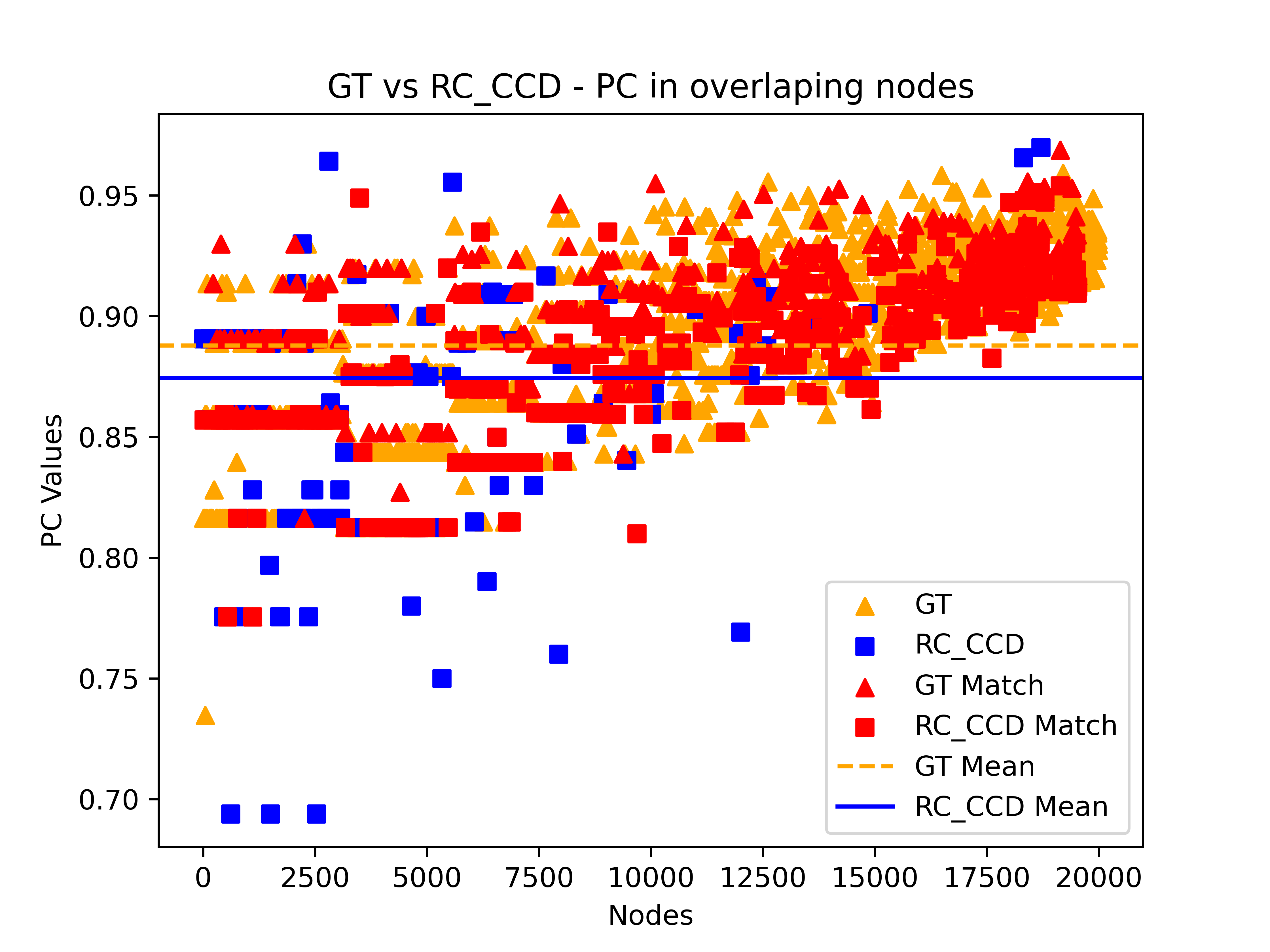}}
\caption{
Comparison of Participation Coefficient values for overlapping nodes identified by RC-CCD and ground-truth communities overlapping nodes on the $LFR$ benchmark.\label{nets_Overlapping_score}}
\end{center}
\end{figure*}

Figure \ref{nets_Overlapping_score} further illustrates these findings for networks with a complexity of \(\mu = 0.6\) (\(net11\)), covering both small and large network sizes. In large networks, nodes identified as overlapping generally showed higher PC values, whereas smaller networks exhibited a more compact distribution of PC values, reflecting the RC-CCD method's precision in simpler network contexts.

\subsection{Boundary Structure Under Different Upper Approximation Scales}
The \(\gamma\) parameter in our RC-CCD method introduces a level of adaptability, allowing for the fine-tuning of overlapping node detection across various network structures. By adjusting this parameter, we will have boundaries of communities more or less tight. In the above experiments, we set the parameter \( \gamma = 0.8\) as a basic level for detecting community structures. To understand how RC-CCD adapts to changes in \( \gamma \), we studied three other values, namely 0.5, 0.6, and 0.7, keeping \( \beta \) constant. \( \gamma \) acts as a tuning parameter with higher values expanding boundaries, and lower values contracting them.

\begin{figure*}[h]
\begin{center}
\subfigure[Small Network ($n=$1000).]{\includegraphics[trim = 1mm 1mm 1mm 8mm, clip, width=65mm]{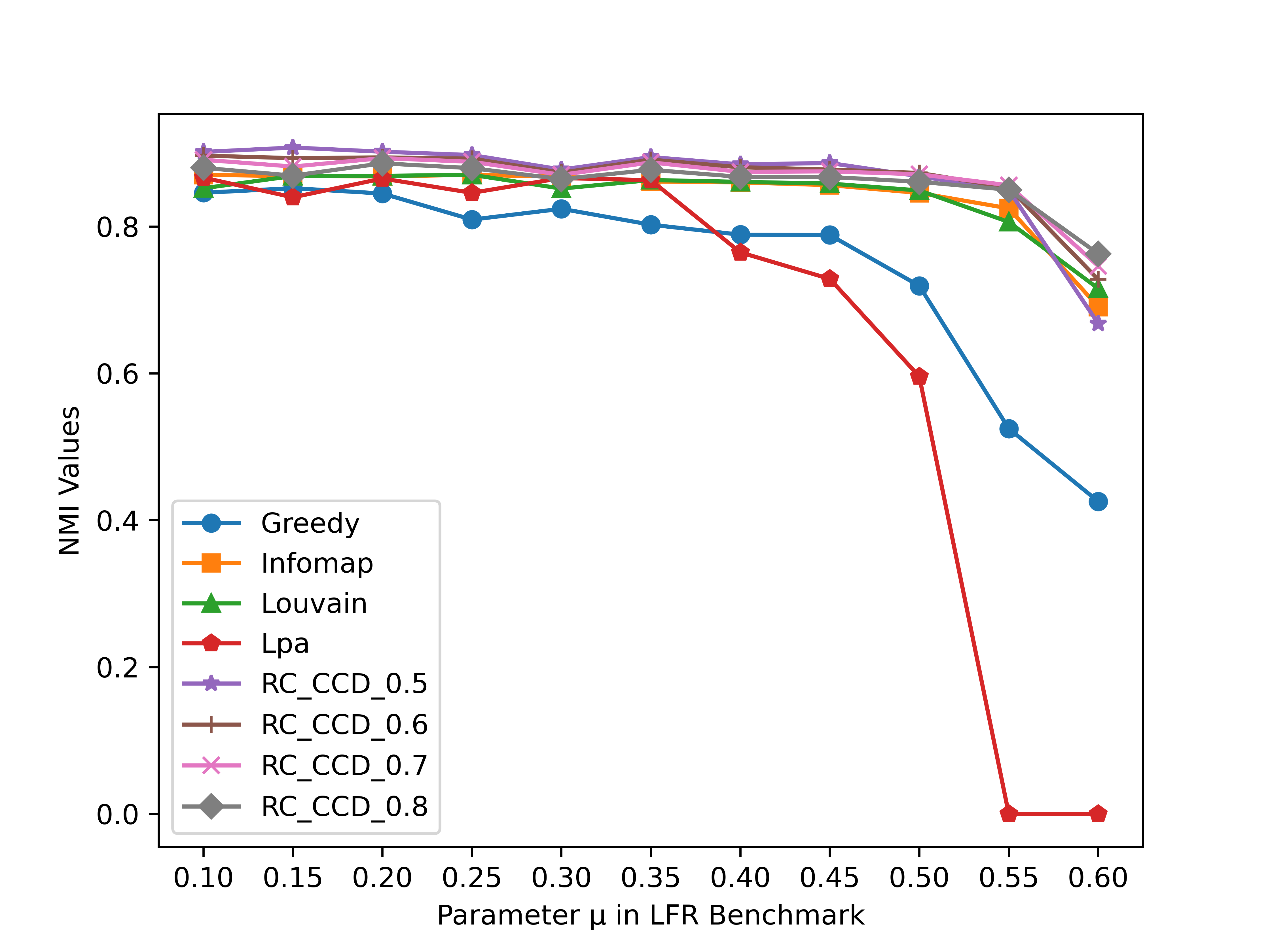}}
\subfigure[Large Network ($n=$20000).]{\includegraphics[trim = 1mm 1mm 1mm 8mm, clip, width=65mm]{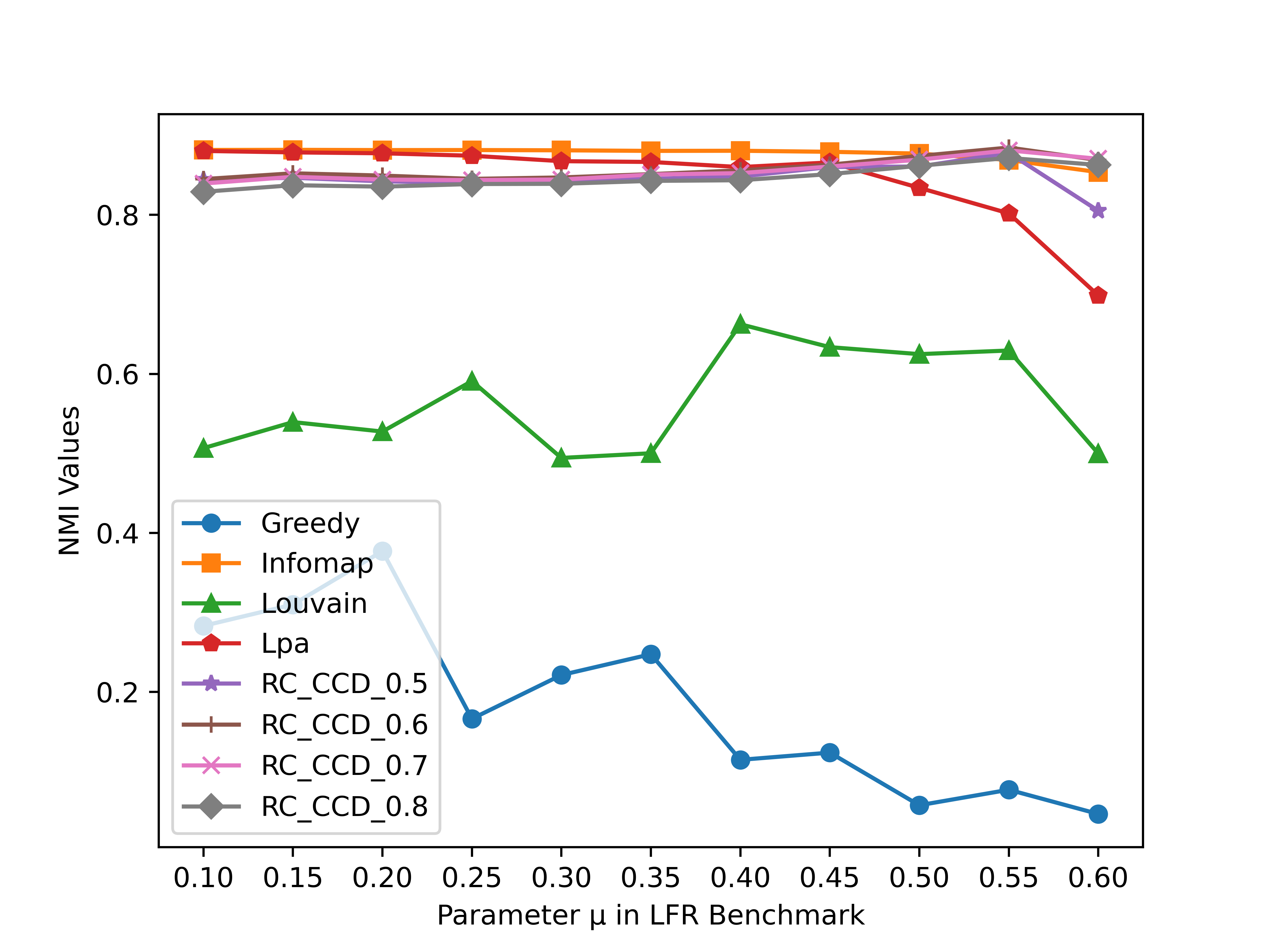}}
\caption{Comparison of base algorithms and RC-CCD (with varying $\gamma$ settings) outcomes on the $LFR$ benchmark.\label{img:nets_NMI_score_gamma}}
\end{center}
\end{figure*}

The effects of varying \( \gamma \) on RC-CCD effectiveness are detailed in Supplementary Tables~\ref{tab:Unified_NMI_values_Small} and \ref{tab:Unified_NMI_values_Large} showing results for NMI in small and large networks, respectively. Figure~\ref{img:nets_NMI_score_gamma} illustrates two examples using the NMI metric in which RC-CCD outperforms all base algorithms under different \( \gamma \) values (refer to Supplementary Table~\ref{tab:Unified_NMI_values_Small} and \ref{tab:Unified_NMI_values_Large} for numeric values). 

Our findings highlight the RC-CCD method's ability to adjust to network complexities via strategic \(\gamma\) parameter tuning. In simpler networks, like \(net1\), a \(\gamma\) setting of 0.5 yielded an optimal NMI score of 0.902, illustrating the advantage of expanding community boundaries in less complex structures. Conversely, in the most intricate network, \(net11\), a higher \(\gamma\) of 0.8 produced the best NMI score of 0.763. This trend is consistent across large networks, where less complex networks achieve higher NMI scores with lower \(\gamma\) values, and the most complex networks reached peak NMI scores with large \(\gamma\) values. Tables~\ref{tab:Unified_NMI_values_Small} and \ref{tab:Unified_NMI_values_Large} corroborate this trend.

In addition to the NMI comparison, Tables~\ref{tab:summaryGammaNet1} and \ref{tab:summaryGammaNet2} present results for Participation Coefficient, for both true and false positives in small and large networks, respectively. RC-CCD performed well when changing the \( \gamma \) values. Note that the results for small networks tended to minimize false positives (FP) across networks of low to medium complexity under various \(\gamma\) values. For networks \(net1\) through \(net7\) specifically, our approach consistently reported zero false positives across all \(\gamma\) settings (0.5, 0.6, and 0.7), accurately identifying true positives (TP) without misclassifying non-membership nodes in simpler networks. In the case of larger networks, the results demonstrate that both true positives and false positives increased with the variation of gamma, being more noticeable with $\gamma$ = 0.5. Overall, RC-CCD proved to be highly effective at identifying overlapping nodes accurately with minimal false positives.

\section{Discussion}\label{sec:discussion}
This work presents RC-CCD, a consensus clustering approach that effectively integrates community structures obtained from multiple detection algorithms to improve precision and adaptability in identifying network community structures. By combining communities from different methodologies, RC-CCD addresses inherent limitations of single-algorithm approaches and leverages their collective strengths, providing a more resilient and versatile solution.

\textbf{Key Findings}: The performance evaluation of RC-CCD using the NMI metric showed a high accuracy in discovering ground-truth communities across networks of varying sizes and complexities. Specifically, RC-CCD achieved an average NMI score of 0.85, significantly surpassing the 0.65 scored by traditional algorithms like Louvain and LPA. This demonstrates the method's robustness and precision in dealing with complex topologies. Moreover, RC-CCD consistently showed over 95\% accuracy in detecting community cores (lower approximations), which are fundamental for determining the number of communities (\(k\)) and building reliable consensus structures.

\textbf{Parameter Sensitivity and Boundary Detection}: A pivotal aspect of our analysis was the exploration of the \( \gamma \) parameter, which fine-tunes the expansion of community boundaries. The results showed that adjusting \( \gamma \) from 0.5 in simpler networks to 0.8 in more complex ones optimized the detection process. Lower \( \gamma \) values facilitated more flexible boundary detection in well-defined networks, while higher \( \gamma \) values improved accuracy in sparse and complex networks. On the other hand, the \( \beta \) parameter provided adaptability by setting the similarity threshold for grouping nodes, offering flexibility based on the diversity of communities generated by the base algorithms.

\textbf{Stability and Overlapping Communities}: Stability analysis revealed that RC-CCD maintained consistent performance across multiple runs, demonstrating its robustness in community detection. This high stability makes RC-CCD a reliable tool for various network domains. Notably, RC-CCD performed very well in detecting overlapping communities, with the mean participation coefficient increasing from 0.65 in simpler networks to 0.89 in more complex ones, closely aligning with the ground truth. This indicates that RC-CCD not only excels in identifying core community structures but also captures nuanced overlapping regions with high precision.

\textbf{Adaptability Across Network Sizes and Complexities}: RC-CCD demonstrated adaptability in both small and large networks, with strong specificity in smaller networks and high accuracy in detecting true positives in more complex networks. The flexibility offered by the method's parameters allowed it to effectively balance between identifying well-defined communities and detecting ambiguous boundary regions.

\textbf{Limitations and Future Work}: While RC-CCD performed robustly across a wide range of synthetic networks, further validation on real-world and dynamic networks is needed to fully assess its versatility. Additionally, fine-tuning the \( \gamma \) and \( \beta \) parameters for specific applications could provide further insights into their optimal ranges, particularly in dynamic networks. Future work could focus on extending RC-CCD to better handle time-varying networks.

\section{Conclusions}
\label{sec:conclusion}
This paper presented RC-CCD, a Rough Set Theory-based consensus clustering framework designed to improve the accuracy, adaptability, and robustness of community detection by integrating community structures from multiple detection algorithms. By leveraging the collective strengths of different algorithms, RC-CCD addresses limitations inherent in single-algorithm approaches, offering a versatile and resilient method for handling diverse and complex networks.

RC-CCD demonstrated a significant improvement in accuracy, achieving an average NMI score of 0.85, which is considerably higher than traditional methods like Louvain and LPA. This highlights RC-CCD’s effectiveness in capturing complex community structures, particularly in networks with high variability in size and complexity. The method also consistently achieved over 95\% accuracy in detecting community cores, which are crucial for determining the overall community structure and ensuring the reliability of the final consensus communities.

The flexibility of RC-CCD is evidenced by its ability to fine-tune boundary detection through the \( \gamma \) parameter. Lower \( \gamma \) values facilitated more flexible boundary identification in well-defined communities, while higher values proved essential for managing sparsely connected and more complex networks. This adaptability makes RC-CCD suitable for a wide range of network applications, enabling precise boundary detection tailored to the network's structure.

RC-CCD’s robustness was further demonstrated through its stability across multiple runs and consistent detection of overlapping communities. The mean participation coefficient improved from 0.65 in simpler networks to 0.89 in more complex ones, confirming RC-CCD’s capability to accurately capture overlapping community regions and complex node relationships.

While RC-CCD has proven effective in handling synthetic networks, several limitations should be acknowledged. First, the model’s performance is sensitive to parameter selection (\( \gamma \) and \( \beta \)), which may require fine-tuning for specific applications. Additionally, real-world networks and dynamic networks present further challenges that need to be addressed. Future research could explore the use of advanced fuzzy logic methods, the application of RC-CCD to evolving and time-varying networks, and the development of automated techniques for optimal parameter selection in diverse network contexts.

\section*{CRediT authorship contribution statement}
\textbf{Darian H. Grass-Boada:} Conceptualization, Methodology, Writing – review \&
editing, original draft. \textbf{Leandro Gonz\'alez-Montesino:} Methodology, Software, Investigation, Writing - original draft. \textbf{Rub\'en Arma\~{n}anzas:} Writing –
review \& editing. 

\section*{Declaration of Competing Interest}
The authors declare that they have no known competing financial interests or personal relationships that could have appeared to influence the work reported in this paper.

\section*{Data availability}
I have shared the links to my data/code where appropriate.

\section*{Acknowledgment}
This work was partially supported by the Gobierno de Navarra through the ANDIA 2021 program (grant no. 0011-3947-2021-000023) and the ERA PerMed JTC2022 PORTRAIT project (grant no. 0011-2750-2022-000000).

\newpage
\appendix

\begin{center}
{\Huge \textbf{Supplementary material}}
\end{center}
\vspace{1cm}

\section{NMI quality measure}
\label{NMIQualityTest}

\begin{table}[H]
\centering
\small
\begin{tabular}{l|cccccc}
\toprule
Nets & RC-CCD & Infomap & Greedy & Lpa & Louvain \\
\midrule
net1 & \textbf{0.880} & 0.870 & 0.846 & 0.867 & 0.852 \\
net2 & \textbf{0.869} & \textbf{0.869} & 0.852 & 0.840 & \textbf{0.869} \\
net3 & \textbf{0.886} & 0.868 & 0.845 & 0.865 & 0.869 \\
net4 & \textbf{0.880} & 0.870 & 0.809 & 0.846 & 0.870 \\
net5 & 0.865 & \textbf{0.868} & 0.824 & 0.866 & 0.852 \\
net6 & \textbf{0.877} & 0.861 & 0.803 & 0.863 & 0.863 \\
net7 & \textbf{0.868} & 0.860 & 0.789 & 0.765 & 0.861 \\
net8 & \textbf{0.868} & 0.856 & 0.789 & 0.729 & 0.859 \\
net9 & \textbf{0.861} & 0.846 & 0.719 & 0.596 & 0.849 \\
net10 & \textbf{0.850} & 0.825 & 0.525 & 0.000 & 0.806 \\
net11 & \textbf{0.763} & 0.691 & 0.426 & 0.000 & 0.716 \\
\bottomrule
\end{tabular}
\caption{NMI values for the proposed method (RC-CCD) and base algorithms across various small networks configuration.}
\label{tab:NMI_Values_SmallNets}
\end{table}

\begin{table}[H]
\centering
\small
\begin{tabular}{l|ccccc}
\toprule
Nets & RC-CCD & Infomap & Greedy & Lpa & Louvain \\
\midrule
net1 & 0.829 & \textbf{0.882} & 0.283 & 0.880 & 0.507 \\
net2 & 0.837 & \textbf{0.882} & 0.310 & 0.879 & 0.539 \\
net3 & 0.836 & \textbf{0.882} & 0.377 & 0.877 & 0.528 \\
net4 & 0.839 & \textbf{0.882} & 0.166 & 0.874 & 0.591 \\
net5 & 0.839 & \textbf{0.881} & 0.221 & 0.868 & 0.494 \\
net6 & 0.843 & \textbf{0.881} & 0.248 & 0.867 & 0.500 \\
net7 & 0.844 & \textbf{0.881} & 0.115 & 0.860 & 0.662 \\
net8 & 0.851 & \textbf{0.879} & 0.124 & 0.866 & 0.634 \\
net9 & 0.862 & \textbf{0.877} & 0.058 & 0.834 & 0.625 \\
net10 & \textbf{0.872} & 0.869 & 0.077 & 0.802 & 0.629 \\
net11 & \textbf{0.863} & 0.854 & 0.047 & 0.699 & 0.500 \\
\bottomrule
\end{tabular}
\caption{NMI values for the proposed method (RC-CCD) and base algorithms across various large networks configuration.}
\label{tab:NMI_Values_LargeNets}
\end{table}

\section{Stability}
\label{StabilityTest}

\begin{sidewaystable}
    \small
    \caption{Small Network Configuration. Summary of Results for Different \( \mu \) Values}
    \label{tab:summarynmismall}
    \resizebox{1.0\textwidth}{!}{%
    \begin{tabular}{|l|c c c |c c c |c c c |}
    \cline{2-10}   
    \multicolumn{1}{c|}{} & \multicolumn{3}{c|}{\( \mu = 0.1 \)} & \multicolumn{3}{c|}{\( \mu = 0.35 \)} & \multicolumn{3}{c|}{\( \mu = 0.6 \)} \\
    \hline
    ALg & 10 & 50 & 100 & 10 & 50 & 100 & 10 & 50 & 100 \\
    \hline
    LPA & 0.87$\pm$0.00 & 0.87$\pm$0.00 & 0.87$\pm$0.00 & 0.87$\pm$0.00 & 0.87$\pm$0.00 & 0.87$\pm$0.00 & 0.00 $\pm$ 0.00 & 0.00 $\pm$ 0.00 & 0.00 $\pm$ 0.00 \\
    Greedy & 0.85$\pm$0.00 & 0.85$\pm$0.00 & 0.85$\pm$0.00 & 0.80$\pm$0.01 & 0.80$\pm$0.01 & 0.80$\pm$0.01 & 0.43 $\pm$ 0.05 & 0.41 $\pm$ 0.05 & 0.43 $\pm$ 0.04 \\
    Louvain & 0.86$\pm$0.01 & 0.86$\pm$0.01 & 0.86$\pm$0.01 & 0.86$\pm$0.01 & 0.86$\pm$0.01 & 0.86$\pm$0.01 & 0.71 $\pm$ 0.03 & 0.70 $\pm$ 0.03 & 0.70 $\pm$ 0.03 \\
    Infomap & 0.87$\pm$0.00 & 0.87$\pm$0.01 & 0.87$\pm$0.00 & 0.86$\pm$0.01 & 0.86$\pm$0.01 & 0.86$\pm$0.01 & 0.67 $\pm$ 0.04 & 0.66 $\pm$ 0.06 & 0.68 $\pm$ 0.07 \\
    RC-CCD & 0.88$\pm$0.00 & 0.88$\pm$0.00 & 0.88$\pm$0.00 & 0.87$\pm$0.00 & 0.87$\pm$0.00 & 0.87$\pm$0.00 & 0.76 $\pm$ 0.01 & 0.73 $\pm$ 0.01 & 0.76 $\pm$ 0.00 \\
    \hline
    \end{tabular}%
    }
    \bigskip \bigskip
    \small
    \caption {Large Network Configuration. Summary of results for different values \( \mu \)}
    \label{tab:summarynmilarge}
    \resizebox{1.0\textwidth}{!}{%
    \begin{tabular}{|l|c c c |c c c |c c c |}
    \cline{2-10}   
    \multicolumn{1}{c|}{} & \multicolumn{3}{c|}{\( \mu = 0.1 \)} & \multicolumn{3}{c|}{\( \mu = 0.35 \)} & \multicolumn{3}{c|}{\( \mu = 0.6 \)} \\
    \hline
    Alg & 10 & 50 & 100 & 10 & 50 & 100 & 10 & 50 & 100 \\
    \hline
    LPA & 0.87$\pm$0.00 & 0.87$\pm$0.00 & 0.87$\pm$0.00 & 0.86$\pm$0.00 & 0.86$\pm$0.00 & 0.86$\pm$0.00 & 0.71$\pm$0.00 & 0.71$\pm$0.00 & 0.69$\pm$0.00 \\
    Greedy & 0.39$\pm$0.07 & 0.39$\pm$0.07 & 0.39$\pm$0.07 & 0.17$\pm$0.00 & 0.17$\pm$0.00 & 0.17$\pm$0.00 & 0.03$\pm$0.00 & 0.03$\pm$0.00 & 0.03$\pm$0.00 \\
    Louvain & 0.50$\pm$0.05 & 0.52$\pm$0.05 & 0.49$\pm$0.05 & 0.57$\pm$0.00 & 0.56$\pm$0.00 & 0.59$\pm$0.00 & 0.63$\pm$0.00 & 0.64$\pm$0.00 & 0.62$\pm$0.00 \\
    Infomap & 0.88$\pm$0.00 & 0.88$\pm$0.04 & 0.88$\pm$0.03 & 0.88$\pm$0.00 & 0.88$\pm$0.00 & 0.88$\pm$0.00 & 0.85$\pm$0.00 & 0.85$\pm$0.00 & 0.85$\pm$0.00 \\
    RC-CCD & 0.83$\pm$0.00 & 0.83$\pm$0.00 & 0.83$\pm$0.00 & 0.84$\pm$0.00 & 0.84$\pm$0.00 & 0.83$\pm$0.00 & 0.86$\pm$0.00 & 0.86$\pm$0.00 & 0.87$\pm$0.00 \\
    \hline
    \end{tabular}%
    }
\end{sidewaystable}

\newpage
\section{Number of Communities}
\label{NumberofK}

\begin{table}[H]
\centering
\small
\begin{tabular}{c c c c c}
\toprule
Net & \multicolumn{2}{c}{GT (k)} & \multicolumn{2}{c}{RC-CCD (k)} \\
\cmidrule(r){2-3} \cmidrule(r){4-5}
    & Small & Large & Small & Large \\
\midrule
net1  & 31 & 337 & 31 & 310 \\
net2  & 31 & 337 & 31 & 310 \\
net3  & 31 & 337 & 31 & 311 \\
net4  & 31 & 337 & 31 & 313 \\
net5  & 31 & 337 & 30 & 314 \\
net6  & 31 & 337 & 31 & 313 \\
net7  & 31 & 337 & 31 & 315 \\
net8  & 31 & 337 & 31 & 318 \\
net9  & 31 & 337 & 31 & 323 \\
net10 & 31 & 337 & 33 & 330 \\
net11 & 31 & 337 & 40 & 366 \\
\bottomrule
\end{tabular}
\caption{Comparison of the Number of Communities (\( k \)) Identified by the Proposed Method and the Ground Truth Across Small and Large Networks}
\label{tab:Community_Count_Comparison}
\end{table}

\section{NMI quality measure with different $\gamma$ values}
\label{NMIQualityDiffGamma}

\begin{table}[H]
\centering
\small
\begin{tabular}{c|cccc|cccc}
\toprule
Nets & \multicolumn{4}{c|}{RC-CCD (NMI, different \(\gamma\))} & \multicolumn{4}{c}{Base Algorithms} \\
\midrule
 & 0.5 & 0.6 & 0.7 & 0.8 & Infomap & Greedy & Lpa & Louvain \\
\midrule
net1  & \textbf{0.902} & 0.897 & 0.891 & 0.880 & 0.870 & 0.846 & 0.867 & 0.852 \\
net2  & \textbf{0.908} & 0.893 & 0.882 & 0.869 & 0.869 & 0.852 & 0.840 & 0.869 \\
net3  & \textbf{0.902} & 0.894 & 0.893 & 0.886 & 0.868 & 0.845 & 0.865 & 0.869 \\
net4  & \textbf{0.897} & 0.892 & 0.888 & 0.880 & 0.870 & 0.809 & 0.846 & 0.870 \\
net5  & 0.878 & 0.873 & 0.870 & 0.865 & \textbf{0.868} & 0.824 & 0.866 & 0.852 \\
net6  & \textbf{0.895} & 0.891 & 0.887 & 0.877 & 0.861 & 0.803 & 0.863 & 0.863 \\
net7  & \textbf{0.885} & 0.881 & 0.875 & 0.868 & 0.860 & 0.789 & 0.765 & 0.861 \\
net8  & \textbf{0.887} & 0.878 & 0.875 & 0.868 & 0.856 & 0.789 & 0.729 & 0.859 \\
net9  & 0.868 & \textbf{0.873} & 0.871 & 0.861 & 0.846 & 0.719 & 0.596 & 0.849 \\
net10 & 0.850 & 0.853 & \textbf{0.856} & 0.850 & 0.825 & 0.525 & 0.000 & 0.806 \\
net11 & 0.668 & 0.728 & 0.746 & \textbf{0.763} & 0.691 & 0.426 & 0.000 & 0.716 \\
\bottomrule
\end{tabular}
\caption{Unified table of NMI values for the proposed RC-CCD method across different values of \(\gamma\) and comparison with base algorithms in the Small Network.}
\label{tab:Unified_NMI_values_Small}
\end{table}

\begin{table}[H]
\centering
\small
\begin{tabular}{c|cccc|cccc}
\toprule
Nets & \multicolumn{4}{c|}{RC-CCD (NMI, different \(\gamma\))} & \multicolumn{4}{c}{Base Algorithms} \\
\midrule
 & 0.5 & 0.6 & 0.7 &    0.8 & Infomap & Greedy & Lpa & Louvain \\
\midrule
net1  & 0.844 & 0.845 & 0.839 & 0.829 & \textbf{0.882} & 0.283 & 0.880 & 0.507 \\
net2  & 0.846 & 0.852 & 0.847 & 0.837 & \textbf{0.882} & 0.310 & 0.879 & 0.539 \\
net3  & 0.842 & 0.849 & 0.844 & 0.836 & \textbf{0.882} & 0.377 & 0.877 & 0.528 \\
net4  & 0.840 & 0.845 & 0.843 & 0.839 & \textbf{0.882} & 0.166 & 0.874 & 0.591 \\
net5  & 0.843 & 0.847 & 0.844 & 0.839 & \textbf{0.881} & 0.221 & 0.868 & 0.494 \\
net6  & 0.845 & 0.851 & 0.850 & 0.843 & \textbf{0.881} & 0.248 & 0.867 & 0.500 \\
net7  & 0.848 & 0.856 & 0.852 & 0.844 & \textbf{0.881} & 0.115 & 0.860 & 0.662 \\
net8  & 0.860 & \textbf{0.863} & 0.860 & 0.851 & 0.879 & 0.124 & 0.866 & 0.634 \\
net9  & 0.860 & \textbf{0.874} & 0.869 & 0.862 & 0.877 & 0.058 & 0.834 & 0.625 \\
net10 & 0.877 & \textbf{0.884} & 0.881 & 0.872 & 0.869 & 0.077 & 0.802 & 0.629 \\
net11 & 0.805 & 0.868 & \textbf{0.870} & 0.863 & 0.854 & 0.047 & 0.699 & 0.500 \\
\bottomrule
\end{tabular}
\caption{Unified table of NMI values for the proposed RC-CCD method across different values of \(\gamma\) and comparison with base algorithms in the Large Network.}
\label{tab:Unified_NMI_values_Large}
\end{table}

\section{Participation coefficient with different $\gamma$ values}
\label{ParticipCoeff}

\begin{sidewaystable}
\small
\caption{Small Network Configuration. \\
Summary of participation coefficient results for different \( \gamma \) Values}
\label{tab:summaryGammaNet1}
\resizebox{0.9\textwidth}{!}{%
\begin{tabular}{|l|c c c c|c c c c|c c c c|}
\cline{2-13}
\multicolumn{1}{c|}{} & \multicolumn{4}{c|}{\( \gamma = 0.5 \)} & \multicolumn{4}{c|}{\( \gamma = 0.6 \)} & \multicolumn{4}{c|}{\( \gamma = 0.7 \)} \\
\hline
Nets & PC\_GT & PC\_RC-CCD & TP & FP & PC\_GT & PC\_RC-CCD & TP & FP & PC\_GT & PC\_RC-CCD & TP & FP \\
\hline
net1  & 0.65 & 0.62 & 26 & 0 & 0.65 & 0.61 & 23 & 0 & 0.65 & 0.60 & 18 & 0 \\
net2  & 0.70 & 0.70 & 34 & 0 & 0.70 & 0.68 & 23 & 0 & 0.70 & 0.68 & 14 & 0 \\
net3  & 0.73 & 0.71 & 28 & 0 & 0.73 & 0.70 & 22 & 0 & 0.73 & 0.69 & 21 & 0 \\
net4  & 0.75 & 0.72 & 24 & 0 & 0.75 & 0.71 & 21 & 0 & 0.75 & 0.71 & 17 & 0 \\
net5  & 0.78 & 0.78 & 22 & 0 & 0.78 & 0.78 & 19 & 0 & 0.78 & 0.77 & 16 & 0 \\
net6  & 0.80 & 0.78 & 28 & 0 & 0.80 & 0.78 & 24 & 0 & 0.80 & 0.78 & 21 & 0 \\
net7  & 0.82 & 0.80 & 22 & 0 & 0.82 & 0.80 & 19 & 0 & 0.82 & 0.79 & 14 & 0 \\
net8  & 0.84 & 0.82 & 31 & 2 & 0.84 & 0.81 & 24 & 2 & 0.84 & 0.82 & 18 & 0 \\
net9  & 0.85 & 0.84 & 37 & 9 & 0.85 & 0.85 & 32 & 5 & 0.85 & 0.84 & 26 & 2 \\
net10 & 0.88 & 0.87 & 37 & 11 & 0.88 & 0.86 & 31 & 9 & 0.88 & 0.85 & 24 & 6 \\
net11 & 0.88 & 0.89 & 44 & 45 & 0.88 & 0.87 & 37 & 34 & 0.88 & 0.86 & 31 & 24 \\
\hline
\end{tabular}%
}

\bigskip \bigskip

\small
\caption{Large Network configuration. \\
Summary of participation coefficient results for different \( \gamma \) Values}
\label{tab:summaryGammaNet2}
\resizebox{0.9\textwidth}{!}{%
\begin{tabular}{|l|c c c c|c c c c|c c c c|}
\cline{2-13}
\multicolumn{1}{c|}{} & \multicolumn{4}{c|}{\( \gamma = 0.5 \)} & \multicolumn{4}{c|}{\( \gamma = 0.6 \)} & \multicolumn{4}{c|}{\( \gamma = 0.7 \)} \\
\hline
Nets & PC\_GT & PC\_RC-CCD & TP & FP & PC\_GT & PC\_RC-CCD & TP & FP & PC\_GT & PC\_RC-CCD & TP & FP \\
\hline
net1  & 0.66 & 0.74 & 638 & 452 & 0.66 & 0.67 & 539 & 104 & 0.66 & 0.65 & 408 & 36\\
net2  & 0.70 & 0.79 & 669 & 638 & 0.70 & 0.73 & 544 & 213 & 0.70 & 0.69 & 419 & 73\\
net3  & 0.73 & 0.81 & 594 & 608 & 0.73 & 0.77 & 484 & 243 & 0.73 & 0.74 & 360 & 102\\
net4  & 0.75 & 0.83 & 382 & 511 & 0.75 & 0.80 & 320 & 296 & 0.75 & 0.76 & 248 & 188\\
net5  & 0.78 & 0.86 & 508 & 503 & 0.78 & 0.82 & 439 & 245 & 0.78 & 0.79 & 322 & 68\\
net6  & 0.80 & 0.88 & 548 & 604 & 0.80 & 0.83 & 471 & 316 & 0.80 & 0.80 & 384 & 142\\
net7  & 0.82 & 0.88 & 568 & 437 & 0.82 & 0.86 & 478 & 182 & 0.82 & 0.84 & 374 & 115\\
net8  & 0.84 & 0.89 & 626 & 368 & 0.84 & 0.85 & 526 & 116 & 0.84 & 0.83 & 414 & 39\\
net9  & 0.86 & 0.91 & 600 & 311 & 0.86 & 0.87 & 531 & 101 & 0.86 & 0.86 & 409 & 55\\
net10 & 0.88 & 0.91 & 623 & 173 & 0.88 & 0.88 & 551 & 153 & 0.88 & 0.88 & 409 & 68\\
net11 & 0.89 & 0.90 & 706 & 362 & 0.89 & 0.88 & 621 & 248 & 0.89 & 0.88 & 462 & 165\\
\hline
\end{tabular}%
}
\end{sidewaystable}

\end{document}